%% file: main.tex
\documentclass[conference]{IEEEtran}
\IEEEoverridecommandlockouts

\usepackage{cite}
\usepackage{url}
\usepackage{amsmath,amssymb,amsfonts}
\usepackage{graphicx}
\usepackage{textcomp}
\usepackage[dvipsnames]{xcolor}
\usepackage{float}
\usepackage{pgffor}
\usepackage{subcaption}
\usepackage{adjustbox}
\usepackage{longtable}
\usepackage{booktabs}
\usepackage{placeins}
\usepackage{algorithm}
\usepackage[noend]{algpseudocode}
\usepackage{ltablex}
\usepackage{array}
\usepackage{siunitx}\usepackage[normalem]{ulem} 
\usepackage{booktabs}
\usepackage[table]{xcolor}
\usepackage{tabularx}
\usepackage{array}
\usepackage{tabularx}
\usepackage{booktabs}
\usepackage{longtable}
\usepackage[table]{xcolor}
\usepackage{pgfplots}
\usepgfplotslibrary{groupplots}

\keepXColumns

\def\BibTeX{{\rm B\kern-.05em{\sc i\kern-.025em b}\kern-.08em
    T\kern-.1667em\lower.7ex\hbox{E}\kern-.125emX}}

\newcommand{\WD}{\textsc{wd}}

\newcommand{\se}[1]{{\fontsize{5.5}{6}\selectfont$\pm$#1}}
\let\reviewStrike\sout
\newcommand{\reviewChange}[1]{{\let\sout\reviewStrike #1}}

\newif\ifrevdraft
\revdraftfalse
\ifrevdraft
  \newcommand{\revTODO}[1]{{\color{red}\textbf{[TODO: #1]}}}
  
\else
  \newcommand{\revTODO}[1]{}
  
  \renewcommand{\sout}[1]{\ignorespaces}
\fi
\renewcommand{\sout}[1]{\ignorespaces}

\begin{document}
\bstctlcite{IEEEexample:BSTcontrol}

\title{WETW: Learning Pair-Specific Weighted Euclidean Distances for Efficient DTW Search}

\author{%
\IEEEauthorblockN{Christos Panourgias\IEEEauthorrefmark{1}, Qitong Wang\IEEEauthorrefmark{2}, George Tzagkarakis\IEEEauthorrefmark{3}, Themis Palpanas\IEEEauthorrefmark{1}}
\IEEEauthorblockA{\IEEEauthorrefmark{1}LIPADE, Universit{\'e} Paris Cit{\'e}, France\quad
\IEEEauthorrefmark{2}Harvard University, Cambridge, MA, USA\quad
\IEEEauthorrefmark{3}FORTH-ICS, Heraklion, Greece \\
panourgiaschris@gmail.com, qitong@seas.harvard.edu, gtzag@ics.forth.gr, themis@mi.parisdescartes.fr}
}

\maketitle

\begin{abstract}\boldmath
Dynamic Time Warping (DTW) is the standard distance for similarity search over time series that are misaligned in time, but its cost is quadratic in the series length and lower bounds often prune few candidates, so search over large collections of long series is slow. Learned approximations embed the series to make comparisons cheap, yet compare every pair with the same fixed rule, usually the Euclidean distance, although DTW aligns each pair differently. We observe that, once a pair is aligned, DTW is a Euclidean distance over coordinates selected for that pair. Building on this view, we propose \underline{W}eighted \underline{E}uclidean \underline{T}ime \underline{W}arping (\textsc{WETW}), a learned distance that adapts the comparison of fixed-length embeddings to each pair. A small symmetric network predicts one weight per embedding dimension and pair without computing an alignment, which halves the error of the best static distance on the same embeddings. The embeddings come from the \underline{W}eighted \underline{E}uclidean \underline{SE}ries \underline{E}ncoder (\textsc{WESee}), whose multi-scale Temporal Feature Extractor keeps the timing information that global pooling removes and raises recall@1 from 0.90 to 0.95, although \textsc{WETW} works with any encoder. We also show that fitting DTW distances is not enough for search, as regression alone misorders the closest neighbours, and that ranking regularization raises recall@1 from 0.79 to 0.95. Our filter-and-verify pipeline, \textsc{WETW-FV}, uses \textsc{WETW} to build a shortlist and exact DTW to check it, so the search is approximate but every returned distance is exact. Over eight datasets, \textsc{WETW-FV} is 11.4$\times$ faster than the fastest exact method among the UCR Suite, lower bounds and MESSI-DTW, at a mean recall@10 of 0.93, while the best learned baseline reaches 0.59 at the same speed. On 48K-point series, \textsc{WETW-FV} remains 40$\times$ faster than the UCR Suite at a recall@10 of 0.91.
\end{abstract}

\begin{IEEEkeywords}
Dynamic Time Warping, Time Series, Similarity Search, Representation Learning
\end{IEEEkeywords}

{\small\noindent\textbf{Code:} {\color{blue}\url{https://github.com/Hendrix8/WETW}}\par}

\section{\textsc{Introduction}}
\label{sec:intro}
\begin{figure}[!t]
 \centering
 \includegraphics{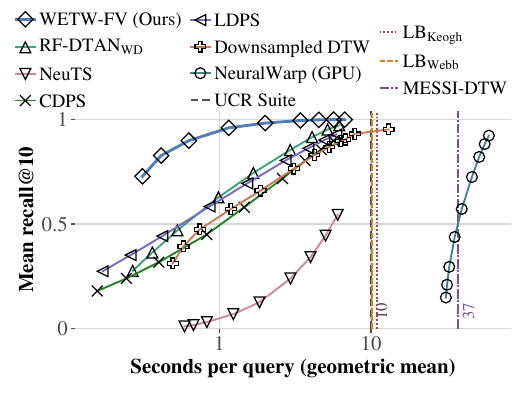}
 \caption{Mean recall@10 versus seconds per query (geometric mean) for 10-NN search over eight datasets. Each curve has nine points, with shortlists from 0.5\% to 50\% of the database.}
 \label{fig:main}

\end{figure}
\begin{figure*}[!t]
 \centering
 \includegraphics[width=.82\linewidth]{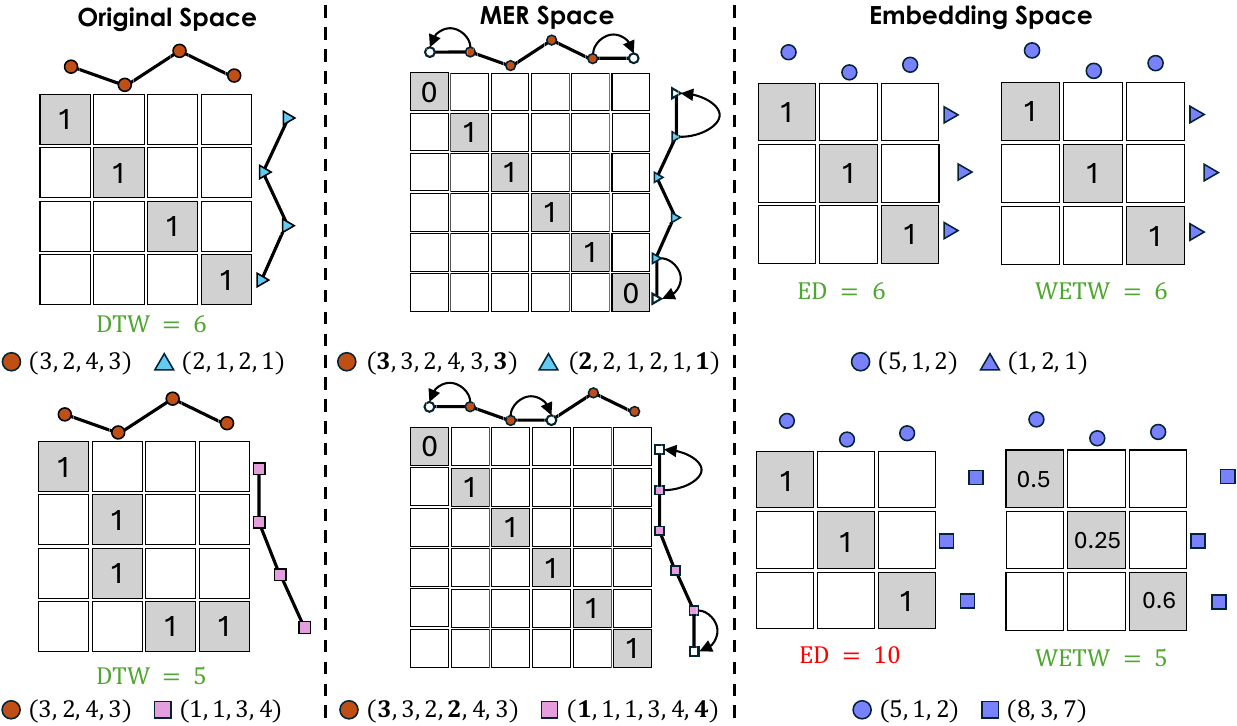}
 \caption{Intuition for pair-specific comparison. DTW paths differ between pairs (left). MER recovers DTW by expanding each pair and masking positions (middle). \textsc{WETW} learns per-pair weights on embedding dimensions, which play a role similar to this mask (right).}
 \label{fig:alignment_to_eucl}
\end{figure*}

\textbf{Similarity search at scale requires efficient DTW comparisons.} Data series are collected at scale in neuroscience, seismology, astrophysics and audio analysis~\cite{Palpanas2015DataSeries, Palpanas2019ITISA}, and similarity search is a basic operation for exploring them~\cite{rakthanmanon2012searching, SEAnet, Kondylakis2018Coconut, Echihabi2021HighDimensional}. A useful comparison must tolerate temporal variation. Dynamic Time Warping (DTW)~\cite{sakoe1978dp} does this by aligning patterns that occur at different speeds or phases~\cite{ratanamahatana2005three,wang2013experimental,bagnall2014experimental,paparrizos2015kshape}, but the alignment is found by dynamic programming, whose cost is quadratic in the series length. On large datasets, repeated DTW computations therefore become the bottleneck of similarity search.


\textbf{Existing accelerations depend on pruning effectiveness and neighbour preservation.} We study whole-series $k$-nearest-neighbour ($k$-NN) search over databases of fixed-length univariate series, where each query is compared with each candidate from start to end, so database representations can be computed once and reused. Exact methods combine lower bounds, early abandoning and indexing to skip unnecessary DTW computations~\cite{rakthanmanon2012searching,11}. However, every candidate that survives the bounds or the index still needs a full DTW computation, and on long series these computations dominate the query time. Uniform downsampling needs no training, but it can smooth away the short events that tell similar series apart. We target medium to long series, where exact pruning works poorly, downsampling loses these neighbours, and DTW remains expensive.

\textbf{Learned filtering reduces the dependence on pruning, downsampling and series length.} Figure~\ref{fig:main} shows these limits on eight datasets. Even the fastest exact search needs about $10$\,s per query (geometric mean). A learned comparison can instead rank the database and pass a shortlist to exact DTW, so every returned distance is exact. Verifying between $2\%$ and $5\%$ of the database, our pipeline is $11.4\times$ faster than the fastest exact search, at a mean recall@10 of $0.93$. At this speed, downsampled DTW reaches $0.50$ and the best learned baseline $0.59$.
\textbf{A fixed comparison rule misses what makes DTW work.} Existing learned approximations of DTW map each series to an embedding and compare embeddings with a fixed distance, usually the Euclidean distance~\cite{ldps,cdps,neuts}, so every pair is compared with the same weights. DTW works differently, since it chooses a new alignment for each pair. Learned embeddings can absorb part of the temporal variation, but a single comparison rule cannot follow alignments that change from pair to pair (Figure~\ref{fig:alignment_to_eucl}, right). This leads to the question we address: \textit{can the comparison itself adapt to each pair, as DTW does, at the cost of a small network?}
\textbf{MER motivates pair-conditioned comparison.} Given an optimal warping path, repeating series elements along the path places matched elements at the same positions, and DTW then equals the squared Euclidean distance between the two expanded series. The expansion has a different length for each pair. To give it a common length, we introduce a \underline{M}aximal \underline{E}xtension \underline{R}epresentation (MER), in which a pair-specific binary mask selects the positions that contribute (Figure~\ref{fig:alignment_to_eucl}, middle, where arrows mark the copied points). In this view, DTW is a Euclidean distance whose coordinates are chosen for each pair. \textsc{WETW} carries this idea over to a learned space. From two independently computed embeddings, and without any warping path, it predicts pair-specific weights that play the role of the MER mask.

\textbf{Our framework approximates DTW without explicit alignment.} We introduce two components: i) the \underline{W}eighted \underline{E}uclidean \underline{SE}ries \underline{E}ncoder (\textsc{WESee}) is a residual convolutional encoder with a multi-scale Temporal Feature Extractor that maps each series to a fixed-size embedding, and ii) \underline{W}eighted \underline{E}uclidean \underline{T}ime \underline{W}arping (\textsc{WETW}) compares two embeddings with a weighted Euclidean distance trained to approximate DTW, in which a small symmetric weight generator sets the weight of each dimension for the pair at hand. \textsc{WETW} only needs embeddings, so it also works with other encoders (Section~\ref{sec:encoders}). For retrieval, our \textsc{WETW} \underline{F}ilter and \underline{V}erify (\textsc{WETW-FV}) pipeline first ranks the stored database embeddings by Euclidean distance, which is cheap and captures part of the temporal variation. \textsc{WETW} then rescores the closest $\kappa\rho N$ of the $N$ series, with $\kappa\geq1$ set on validation, and exact DTW verifies the best $\rho N$ of them.

Our main contributions can be summarized as follows.


\begin{itemize}
\item We introduce \textsc{WESee}, an encoder that maps each series to a $d$-dimensional embedding that can be stored and reused. 
In contrast to existing DTW-preserving encoders (that use global pooling), the multi-scale Temporal Feature Extractor of \textsc{WESee} keeps track of when patterns occur.

\item We show that fitting DTW distances does not preserve the order of the closest neighbours, which is what search depends on. 
To address this problem, we introduce in \textsc{WESee} a ranking regularization term that penalizes misordered neighbours alongside the regression on DTW targets. This term is independent of the encoder and of the distance, so its utility goes beyond \textsc{WESee}: it applies to any learned approximation of DTW.

\item We propose \textsc{WETW}, which (to the best of our knowledge) is the first learned DTW approximation whose distance between stored embeddings adapts to each pair, as DTW itself does (in contrast, existing methods compare embeddings with a static distance, usually the Euclidean one). 
Following the MER view, which translates DTW to Euclidean distance over coordinates selected for each pair, a small symmetric generator predicts one weight per embedding coordinate and pair, without computing any alignment.
Unlike pairwise models that revisit the series or their per-timestep features, \textsc{WETW} only reads two stored embeddings, remains non-negative, symmetric and zero on identical embeddings, and works with any encoder.

\item We design \textsc{WETW-FV}, a filter-and-verify search pipeline in which a cheap Euclidean prefilter selects candidates, \textsc{WETW} rescores the closest of them, and exact DTW verifies a small shortlist, so the search is approximate but every returned distance is exact. 
We analyze its complexity and show how the amount of rescoring trades accuracy for time.

\item We evaluate our approach on eight datasets with series of up to 48K points and databases of up to 840K series. 
Against exact search baselines, \textsc{WETW-FV} is $11.4\times$ faster than the fastest competitor 
at a mean recall@10 of $0.93$, and $40\times$ faster than the fastest competitor 
on long (48K-point) series at a recall@10 of $0.91$. 
The best learned baseline reachs only a mean recall@10 of $0.59$ at the same speed as our method, and recall@10 of $0.60$ on the long series.
On the same embeddings, WETW halves the error of the best static distance ($0.04$ against $0.08$), ranking regularization raises recall@1 from $0.79$ to $0.95$, and the Temporal Feature Extractor raises it from $0.90$ to $0.95$ at matched capacity.
\end{itemize}

\section{\textsc{Related Work}}
\label{sec:related}
\textbf{Main directions for efficient DTW approximation.} Work on efficient DTW-based comparison falls into five groups: (i)~reducing the resolution of the input series~\cite{keogh2001dimensionality,lin2003experiencing}, (ii)~accelerating exact DTW with lower bounds and indexes~\cite{keogh2005exact,rakthanmanon2012searching}, (iii)~DTW-preserving representations~\cite{ldps,cdps}, (iv)~neural similarity learning that approximates expensive distances in embedding spaces~\cite{neuts,grabocka2018neuralwarp}, and (v)~learned temporal alignment that explicitly models warping transformations~\cite{ShapiraWeber2019,rfdtan}. Data series indexes such as iSAX, DSTree and MESSI support exact and approximate similarity search over very large collections, mostly under the Euclidean distance~\cite{isax,wangdataadaptive,Peng2020,lernaean,echihabi2019hydra,gogolou2022pros,levchenko2020bestneighbor}. MESSI also supports DTW~\cite{11}, and we use it as a baseline. Our approach separates representation from comparison. \textsc{WESee} learns fixed-length embeddings, and \textsc{WETW} compares them with a small pair-specific weighted distance. Because \textsc{WETW} only operates on embeddings, it also applies to other encoders.

\textbf{Reducing resolution before exact DTW.} A simple and often overlooked option is to shorten the series before running DTW, by uniform downsampling, piecewise aggregate approximation~\cite{keogh2001dimensionality} or its symbolic version~\cite{lin2003experiencing}. With a fractional warping window radius $w$, constrained DTW costs $\mathcal{O}(wL^2)$, so keeping every $g$-th point cuts the dynamic-programming work by about $g^2$, without any training. On oversampled data this keeps the coarse similarity structure even for large $g$, which makes it a strong baseline. Skipping samples, however, can miss fast changes in the signal, which distorts DTW distances and, for larger $g$, changes the nearest neighbours. Its cost also remains quadratic in $L$ for a fixed $g$. \textsc{WETW} is useful when this loss of detail matters. It learns from full-resolution series, and its ranking cost does not depend on $L$.

\textbf{Exact DTW acceleration with lower bounds.} Exact methods avoid most DTW computations instead of approximating DTW. A cheap lower bound on DTW~\cite{yi1998efficient,keogh2005exact,lemire2009faster,tan2019elastic,webb2021tight} is computed for each candidate, and the candidate is skipped if the bound already exceeds the distance of the best match so far. The remaining candidates are checked with a DTW computation that stops as soon as it exceeds that distance. The UCR Suite combines these steps in a cascade and has made exact DTW search practical~\cite{rakthanmanon2012searching}, and indexes built on such bounds extend it to large databases of whole series~\cite{niennattrakul2009exact,tan2017indexing}. This works well when the bounds are tight and the true matches are clearly closer than the rest of the data, and it fails when either condition does not hold. On oscillating signals such as audio, whale calls or EEG, the signal covers most of its range within each warping window, so the bounds are close to zero. When most candidates lie at similar distances from the query, even a tight bound cannot rule them out. In both cases most candidates still need a full DTW computation. \textsc{WETW} replaces the lower bound with a learned comparison of stored embeddings, whose cost does not depend on how well the bounds prune, and runs exact DTW only on a small shortlist.
\textbf{Distance-preserving representations and neural metric learning.} Several methods map series to a Euclidean space that preserves their DTW similarity. Lods et al.~\cite{ldps} learn shapelet-based representations, \textit{Learning DTW-Preserving Shapelets} (LDPS), from pairwise DTW targets, so that their Euclidean distances approximate DTW. El Amouri et al.~\cite{cdps} extend this idea with \textit{Constrained DTW-Preserving Shapelets} (CDPS), which add must-link and cannot-link constraints to improve clustering. Yao et al.~\cite{neuts} propose \textit{NeuTS}, which approximates several distances, including DTW, with deep metric learning, training an attention-augmented RNN encoder on sampled representative series with a distance-weighted ranking loss. Like \textsc{WETW}, these methods replace expensive distance computations with comparisons in an embedding space, but once the series are embedded, every pair is compared with the same fixed measure. \textsc{WETW} instead computes its weights from both embeddings, so the comparison changes with the pair, which brings the learned distance closer to DTW.


\begin{figure*}[!t]
 \centering

 \begin{minipage}[b]{0.695\textwidth}\centering{\small\bfseries \textsc{WESee}}\end{minipage}\hfill
 \begin{minipage}[b]{0.2696\textwidth}\centering{\small\bfseries \textsc{WETW}}\end{minipage}\\[2pt]
 \begin{minipage}[c]{0.695\textwidth}
 \centering
\includegraphics[width=0.95\linewidth]{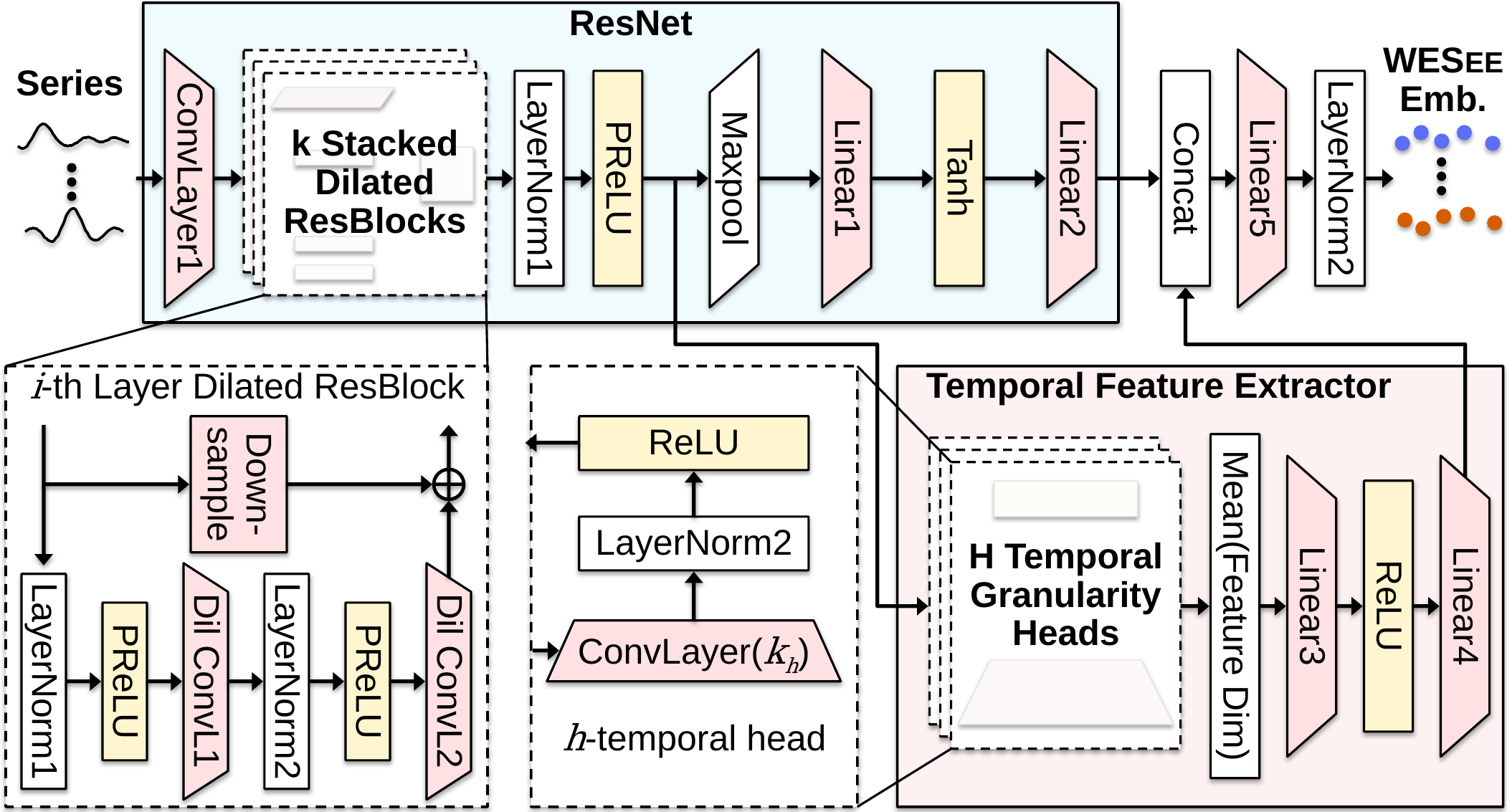}
 \end{minipage}
 \hfill
 \raisebox{\dimexpr-0.1873\textwidth+0.5ex\relax}{%
 \textcolor{black!45}{\rule{0.7pt}{0.3746\textwidth}}}
 \hfill
 \begin{minipage}[c]{0.2696\textwidth}
 \centering
\includegraphics[width=0.95\linewidth]{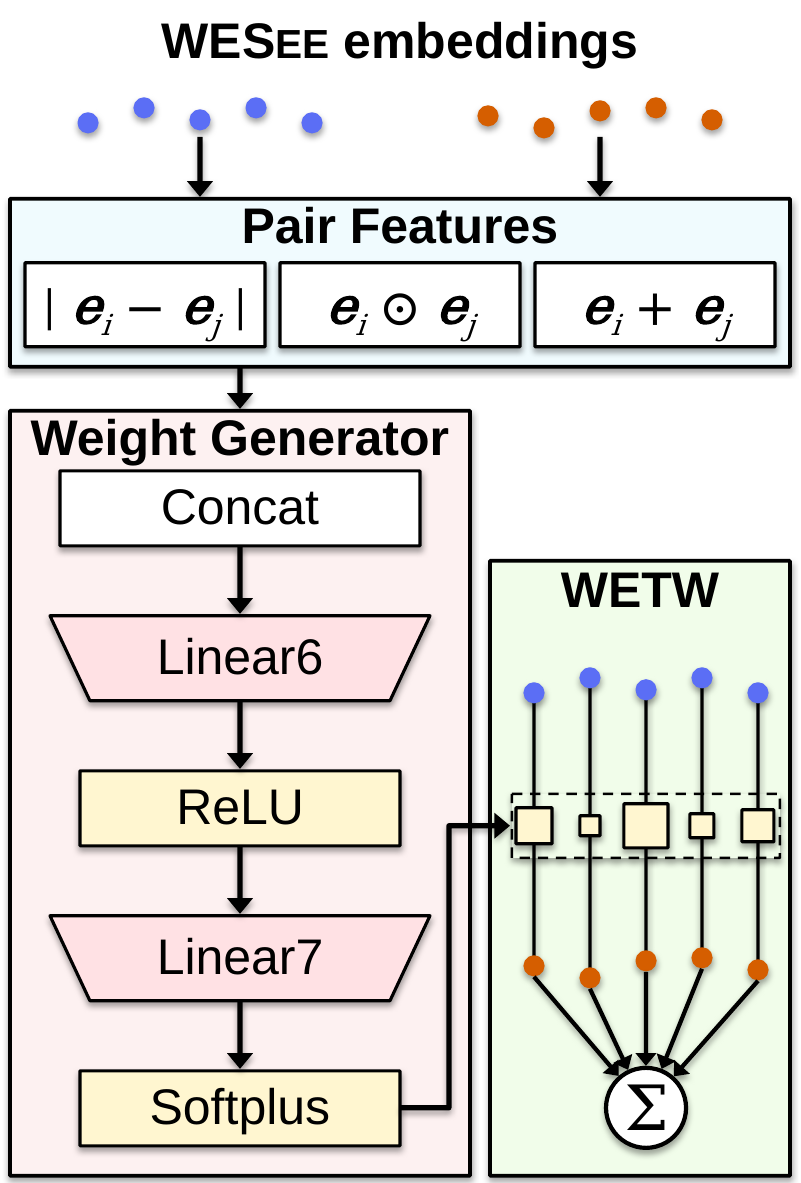}
 \end{minipage}

 \caption{%
(Left) \textsc{WESee} encoder architecture. (Right) WETW architecture.}
 \label{fig:wesee_combined}
\end{figure*}

\textbf{Learning pair-specific distances.} The closest work to ours is NeuralWarp~\cite{grabocka2018neuralwarp}, which also replaces a static metric with a learned \emph{pairwise} function. An encoder maps each series to per-timestep context vectors, and an MLP applied to every pair of time indices predicts a soft alignment weight for the distance between the two contexts. NeuralWarp solves a different problem, however. It is trained from class labels to maximize nearest-neighbour classification accuracy rather than to approximate DTW, so it needs labelled data. Its warping network also runs on all $L^2$ index pairs, so one comparison costs $\mathcal{O}(L^2 d_{\mathrm{ctx}})$ for context vectors of dimension $d_{\mathrm{ctx}}$, which keeps the quadratic dependence on series length. \textsc{WETW} learns to approximate DTW without labels, and its weights act on the $d$ dimensions of fixed-size embeddings, so one comparison costs one generator pass and $\mathcal{O}(d)$ distance operations.
\textbf{Learning temporal alignment.} Other work learns the alignment itself. Shapira Weber et al.~\cite{ShapiraWeber2019} introduce the \textit{Diffeomorphic Temporal Alignment Network} (DTAN), which predicts input-dependent diffeomorphic warping functions in a single forward pass, and its regularization-free version~\cite{rfdtan} replaces hand-crafted warp penalties with a forward--backward consistency constraint. These methods learn the warping directly. \textsc{WETW} instead approximates DTW in an embedding space without computing a warping path or aligning the series.

\textbf{Positioning of \textsc{WETW}.}
\textsc{WETW} is meant as a DTW surrogate for long series with irregular temporal structure, whose DTW distances can differ substantially from their Euclidean distances. For series lengths from 1K to 48K, \textsc{WETW-FV} keeps the highest recall@10 among the learned filters (Section~\ref{sec:main_results}). Our encoder, \textsc{WESee}, builds on the residual backbone of SEAnet~\cite{SEAnet} but has no decoder, which avoids a reconstruction objective that could work against DTW approximation. We also add a multi-scale Temporal Feature Extractor, similar in spirit to InceptionTime~\cite{fawaz2020inceptiontime}, to keep the temporal structure that pooling removes (Section~\ref{sec:WESee}).

\section{Background and Problem Formulation}

\textbf{Data series and DTW.} A univariate data series of length $n$ is an ordered sequence $\mathbf{x}=(x_1,x_2,\ldots,x_n)\in\mathbb{R}^{n}$. For two series $\mathbf{x}\in\mathbb{R}^{n}$ and $\mathbf{y}\in\mathbb{R}^{m}$, DTW finds the warping path with the lowest alignment cost~\cite{sakoe1978dp}. A warping path $\pi=((i_1,j_1),\ldots,(i_K,j_K))$ starts at $(1,1)$, ends at $(n,m)$ and moves by steps in $\{(1,0),(0,1),(1,1)\}$, so it is monotone. We use the accumulated squared cost
{\small
\[
\mathrm{DTW}(\mathbf{x},\mathbf{y})=\min_{\pi \in \mathcal{A}(\mathbf{x},\mathbf{y})}\sum_{(i,j)\in \pi} c(x_i,y_j),
\]
}
where $\mathcal{A}(\mathbf{x},\mathbf{y})$ is the set of valid warping paths and $c(x_i,y_j)=(x_i-y_j)^2$. Dynamic programming over the \(n \times m\) cost matrix computes this minimum in \(\mathcal{O}(nm)\) time.

\textbf{Problem formulation.} We consider whole-series matching, where the query is compared from start to end with each database series. Given a database $\mathcal{D}=\{\mathbf{x}_1,\ldots,\mathbf{x}_N\}$ of univariate series with $\mathbf{x}_i\in\mathbb{R}^{L}$, we learn an encoder $\phi_\theta\colon\mathbb{R}^{L}\to\mathbb{R}^{d}$ and a comparison of embeddings $d_\psi\colon\mathbb{R}^{d}\times\mathbb{R}^{d}\to\mathbb{R}_{\geq0}$, where the embedding dimension $d$ does not depend on $L$. We use Sakoe--Chiba DTW~\cite{sakoe1978dp} with warping window radius $r=\lfloor wL\rfloor$ and $w=0.1$, which restricts paths to $|i-j|\leq r$. Our predictions approximate the length-normalized target
{\small
\[
d_{ij}=\sqrt{\frac{\mathrm{DTW}(\mathbf{x}_i,\mathbf{x}_j)}{L}},
\quad\text{with}\quad
\hat d_{ij}=d_\psi(\phi_\theta(\mathbf{x}_i),\phi_\theta(\mathbf{x}_j)).
\]
}
This monotone transformation preserves DTW neighbour rankings for a fixed $L$, and our goal is to preserve these rankings with a cheaper comparison of stored embeddings. In our framework, $\phi_\theta$ is the \textsc{WESee} encoder and $d_\psi$ is the \textsc{WETW} comparison. Section~\ref{sec:complexity} analyses the cost of each stage and of the whole search relative to exact DTW.

\section{The \textsc{WETW} Distance and the \textsc{WESee} Encoder}

\subsection{Weighted Euclidean Series Encoder \textsc{(WESee)}}
\label{sec:WESee}

\textbf{Global and temporal summaries.} \textsc{WESee} maps each series to a $d$-dimensional embedding meant to keep the temporal structure that matters for DTW. The encoder stacks one-dimensional residual~\cite{He2015} convolutional blocks with exponentially increasing dilation rates~\cite{yu2016multi}, which capture both local motifs and longer-range dependencies while the depth grows only linearly. The result is a latent feature map $\mathbf{T}\in\mathbb{R}^{C\times L'}$ for each series, with $C$ convolutional channels and temporal length $L'$. The residual backbone summarizes \(\mathbf{T}\) by global max pooling over time followed by a projection MLP, which gives a global embedding \(\mathbf{e}_{\mathrm{G}}\in\mathbb{R}^{d}\). This embedding records which patterns a series contains, but not when they occur. The \emph{Temporal Feature Extractor} (TFE) branches off \(\mathbf{T}\) before pooling to keep this timing information. Similar in spirit to InceptionTime~\cite{fawaz2020inceptiontime}, it applies several temporal convolutional heads with different kernel sizes to extract patterns at different time scales. Their outputs are concatenated along the channel dimension and fused, and averaging over channels leaves a temporal profile that an MLP projects to \(\mathbf{e}_{\mathrm{TFE}}\in\mathbb{R}^{d}\). The final embedding is
\[
\mathbf{e}_i=\mathrm{LayerNorm}\!\left(f_{\mathrm{merge}}([\mathbf{e}_{\mathrm{G},i};\mathbf{e}_{\mathrm{TFE},i}])\right),
\]
where $\mathbf{e}_{\mathrm{G},i}$ and $\mathbf{e}_{\mathrm{TFE},i}$ are the global and temporal embeddings of series $i$, $[\,\cdot\,;\,\cdot\,]$ denotes concatenation, and $f_{\mathrm{merge}}\colon\mathbb{R}^{2d}\to\mathbb{R}^{d}$ is a linear projection. The final LayerNorm~\cite{Ba2016LayerN} centres each embedding and normalizes the variance of its coordinates, so that all embeddings are on a common scale. Figure~\ref{fig:wesee_combined} (left) shows the architecture.

\subsection{Weighted Euclidean Time Warping \textsc{(WETW)}}
\label{sec:WETW}

\textbf{From MER to a learned space.} In the MER view of exact DTW (Section~\ref{sec:intro}), a pair-specific expansion puts matched series elements at the same coordinates, and a binary mask selects which coordinates count. This suggests learning a space in which series are compared coordinate by coordinate, together with a pair-specific way to adjust how much each coordinate contributes. \textsc{WESee} maps each series on its own to a shared $d$-dimensional space, and the weights of \textsc{WETW} play a role similar to the MER mask. As in MER, the representation and the comparison rule depend on each other, so we train the encoder and the weight generator jointly. Each embedding coordinate mixes information from several time regions and scales, and DTW supervision can teach the encoder to absorb part of the temporal variation. The remaining gap between embedding distance and DTW still varies from pair to pair, which is why we keep pair-specific weights even after the encoder has learned from DTW. These weights need not correspond to DTW alignments, as they are learned only to make the weighted distance match DTW.

\textbf{Weight generator.}
For embeddings $\mathbf{e}_i=\phi_\theta(\mathbf{x}_i)$ and $\mathbf{e}_j=\phi_\theta(\mathbf{x}_j)$, the generator $\omega_\psi$ outputs %
positive weights $\mathbf{w}_{ij}\in\mathbb{R}^{d}_{>0}$, one per embedding dimension.
It is a two-layer MLP over \emph{symmetric} pair features, and the distance is a scaled weighted Euclidean distance,
\[
\mathbf{w}_{ij}=\mathrm{softplus}\!\big(\mathrm{MLP}([\,|\mathbf{e}_i-\mathbf{e}_j|;\ \mathbf{e}_i\odot\mathbf{e}_j;\ \mathbf{e}_i+\mathbf{e}_j\,])\big),
\]
\[
\hat d_{ij}=s\,\sqrt{\textstyle\sum_{k=1}^{d} w_{ij,k}(e_{i,k}-e_{j,k})^2},
\]
with a learned scale $s>0$. Here $\odot$ is element-wise multiplication, absolute values are taken element-wise, and $w_{ij,k}$ and $e_{i,k}$ are coordinate $k$ of $\mathbf{w}_{ij}$ and $\mathbf{e}_i$.
The absolute difference describes coordinate-wise discrepancies, the element-wise product captures interactions between matching coordinates, and the sum tells the generator where the pair lies in the embedding space. The generator can therefore condition its weights both on the differences and on the embeddings themselves. All three features are unchanged when the two inputs are swapped, so the weights and the distance are symmetric by construction, that is, $\hat d_{ij}=\hat d_{ji}$. Softplus keeps every weight strictly positive, so each dimension adds a non-negative term to the squared distance, and $\hat{d}_{ij}=0$ if and only if the two embeddings are equal. Like DTW, \textsc{WETW} does not necessarily satisfy the triangle inequality. The scale $s$ matches the predictions to the scale of the DTW target. We keep the MLP small because it runs once for every candidate pair, so its cost adds directly to the query time.

\subsection{\textsc{WETW} Filter and Verify \textsc{(WETW-FV)}}
\label{sec:wetw_fv}

\begin{figure}[!t]
 \centering
 \includegraphics[width=0.85\columnwidth]{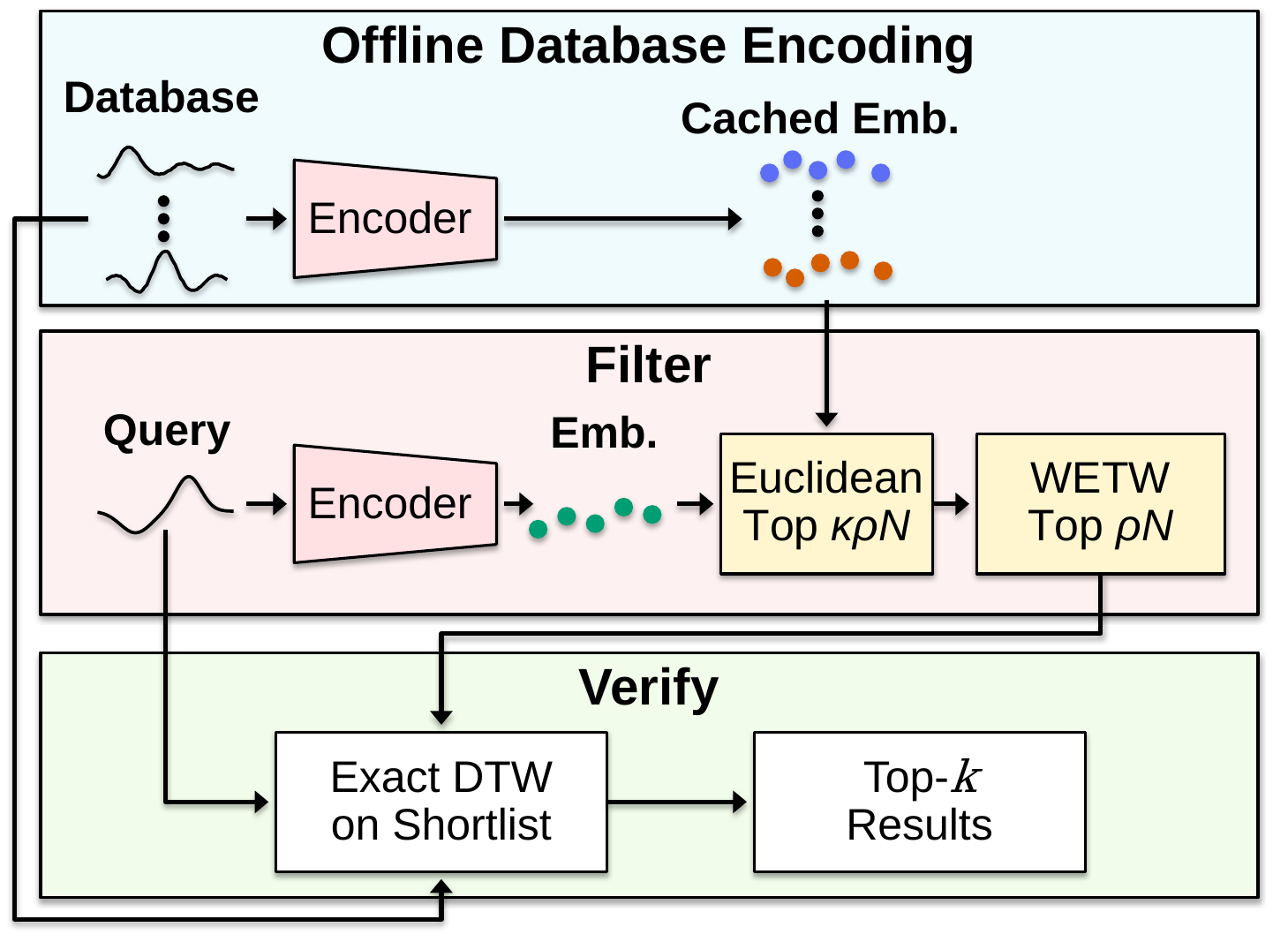}
 \caption{\textsc{WETW-FV} pipeline. A Euclidean prefilter selects candidates, \textsc{WETW} rescores them into a shortlist, and exact DTW on the original series returns the top-$k$ results.}
 \label{fig:wetw_fv}
\end{figure}

\textbf{Euclidean prefiltering.}
\textsc{WETW-FV} uses the learned distance to shortlist candidates and exact DTW to find the final nearest neighbours (Figure~\ref{fig:wetw_fv}). We encode the database series once and store the embeddings. For each query $\mathbf{q}$ we compute $\mathbf{e}_q=\phi_\theta(\mathbf{q})$ and rank the database in two stages. First, a prefilter ranks all stored embeddings by their Euclidean distance to $\mathbf{e}_q$, which is cheap and exploits the partial alignment information already encoded in the embeddings. Second, \textsc{WETW} rescores only the closest candidates. This limits the cost of the generator and focuses \textsc{WETW} on ordering the most promising candidates, where errors in the estimated distances decide which neighbours enter the final shortlist.

\textbf{Shortlist and rescoring sizes.}
The final shortlist holds a fraction $\rho\in(0,1]$ of the $N$ database series, namely the $\rho N$ ($\geq k$) candidates with the smallest \textsc{WETW} distances. \textsc{WETW} rescores $\kappa$ times as many candidates, $\kappa\rho N$ with $\kappa\geq1$, so a larger $\kappa$ depends less on the Euclidean ranking. The ratio $\kappa$ matters most when exact verification is cheap and scoring dominates the query time. We choose $\kappa$ for each $\rho$ on validation data (Section~\ref{sec:exp_setup}), and allow \textsc{WETW} to rescore the whole database when Euclidean prefiltering cannot reduce the scoring cost at similar recall.

\textbf{Exact verification.}
We verify the shortlist with exact DTW on the original series, with the same warping window as the target search, and return the $k$ candidates with the smallest exact distances. Lower bounds and early abandoning reduce the computational burden, and every returned distance is exact. A true nearest neighbour is missed only if the filter excludes it, so a larger $\rho$ can raise recall at the cost of more verification, and $\rho=1$ gives exact search plus the overhead of encoding, scoring and sorting. Algorithm~\ref{alg:wetw_fv} summarizes preprocessing, training and search.

\algrenewcommand\algorithmicrequire{\textbf{Input:}}
\algrenewcommand\algorithmicensure{\textbf{Output:}}
\begin{algorithm}[!t]
\caption{\textsc{WETW}: preprocessing, training, and search}
\label{alg:wetw_fv}
\small
\begin{algorithmic}[1]
\Statex \makebox[\linewidth]{\hrulefill\quad\textbf{1. Preprocessing}\quad\hrulefill}
\Require Disjoint sets $\mathcal{S},\mathcal{V}$; batch size $b_s$; DTW radius $r$
\Ensure Normalized $\mathcal{S},\mathcal{V}$; target caches $\mathcal{C}_{\mathcal S},\mathcal{C}_{\mathcal V}$
\State $\mathbf{x}\gets\operatorname{znorm}(\mathbf{x})$, $\forall\mathbf{x}\in\mathcal{S}\cup\mathcal{V}$
\For{$\mathcal U\in\{\mathcal{S},\mathcal{V}\}$}
 \State Sample pairs $\mathcal P_{\mathcal U}$ and ranking anchors $\mathcal A_{\mathcal U}\subseteq\mathcal U$
 \State $d_{ij}\gets\sqrt{\mathrm{DTW}_r(\mathbf{x}_i,\mathbf{x}_j)/L}$
 \State $\mathcal C_{\mathcal U}\gets\{(i,j,d_{ij}):(i,j)\in\mathcal P_{\mathcal U}\cup(\mathcal A_{\mathcal U}\times\mathcal U)\}$
\EndFor
\State Extend $\mathcal C_{\mathcal S}$ as new training pairs are sampled
\Statex \makebox[\linewidth]{\hrulefill\quad\textbf{2. Training}\quad\hrulefill}
\Require $\mathcal S,\mathcal V,\mathcal C_{\mathcal S},\mathcal C_{\mathcal V}$; $\phi_\theta,\omega_\psi,s$; shortlist fraction $\rho$
\Ensure Trained $\theta,\psi,s$; prefilter ratio $\kappa$
\State $\mathcal L_{\mathrm{MSE}}\gets\frac{1}{|\mathcal P|}\sum_{(i,j)\in\mathcal P}(\|\mathbf e_i-\mathbf e_j\|_2/\sqrt d-d_{ij})^2$
\State $\theta\gets\operatorname{Fit}_{\theta}(\mathcal L_{\mathrm{MSE}};\mathcal S,\mathcal V)$ \Comment{warmup}
\State $(\theta,\psi,s)\gets\operatorname{Fit}_{\theta,\psi,s}(\mathcal L_{\mathrm{WD}};\mathcal S,\mathcal V)$ \Comment{joint}
\State $\theta\gets\operatorname{Fit}_{\theta}(\beta\mathcal L_{\mathrm{WD}}+\mathcal L_{\mathrm{rank}};\mathcal S,\mathcal V)$, Euclidean $\hat d$ \Comment{rank}
\State Freeze $\theta$; cache $\mathbf e_i\gets\phi_\theta(\mathbf x_i)$ for $\mathbf x_i\in\mathcal S\cup\mathcal V$
\State Sample regression pairs $\mathcal P$ and ranking triplets $(a,p,n)$
\State $(\psi,s)\gets\operatorname{Fit}_{\psi,s}(\beta\mathcal L_{\mathrm{WD}}+\mathcal L_{\mathrm{rank}};\mathcal S,\mathcal V)$
\State $\kappa\gets\operatorname{SelectPrefilter}(\mathcal V,\rho)$
\Statex \makebox[\linewidth]{\hrulefill\quad\textbf{3. \textsc{WETW-FV} search}\quad\hrulefill}
\Require $\phi_\theta,\omega_\psi,s$; database $\mathcal D$, $N=|\mathcal D|$; query $\mathbf q$; $k,\rho,\kappa,r$
\Ensure $k$ verified neighbours and exact DTW distances
\State Cache $\mathbf x_j\gets\operatorname{znorm}(\mathbf x_j)$, $\mathbf e_j\gets\phi_\theta(\mathbf x_j)$, $\forall j$
\State $\mathbf q\gets\operatorname{znorm}(\mathbf q)$; $\mathbf e_q\gets\phi_\theta(\mathbf q)$
\State $m\gets\min\{N,\max\{k,\lfloor\rho N\rfloor\}\}$
\State $p\gets\min\{N,\kappa m\}$ \Comment{\textsc{WETW} rescores $\kappa\rho N$}
\State $C\gets\operatorname{Top}_p\{(j,\|\mathbf e_q-\mathbf e_j\|_2):j=1,\ldots,N\}$
\State $\hat d_{qj}\gets d_\psi(\mathbf e_q,\mathbf e_j)$, $\forall j\in C$
\State $S\gets\operatorname{Top}_m\{(j,\hat d_{qj}):j\in C\}$
\State $\delta_j\gets\sqrt{\mathrm{DTW}_r(\mathbf q,\mathbf x_j)/L}$, $\forall j\in S$
\State \Return $\operatorname{Top}_k\{(j,\delta_j):j\in S\}$ and distances
\end{algorithmic}
\end{algorithm}

\subsection{Computational and Storage Complexity}
\label{sec:complexity}

\textbf{\textsc{WESee} encodes each series once.}
Let $L$, $d$ and $N$ be the series length, the embedding dimension and the database size. Sakoe--Chiba DTW with warping window radius $r$ costs $\mathcal{O}(L(r+1))$ per pair in the worst case, which is $\mathcal{O}(L^2)$ for our choice $r=\lfloor0.1L\rfloor$.
Let $T_{\mathrm{enc}}(L)$ be the cost of encoding one series. With $B$ residual blocks, at most $C$ channels and fixed kernel sizes, the backbone costs $\mathcal{O}(BLC^2)$, since at each time step a convolution maps up to $C$ input channels to up to $C$ output channels. The temporal heads and projection layers add work, but with fixed widths, kernel sizes and number of heads, the total encoding cost grows linearly with $L$. This cost is paid once offline for each database series and once at search time for each query. Encoding the database therefore costs $\mathcal{O}(N T_{\mathrm{enc}}(L))$, and the stored embeddings take $\mathcal{O}(Nd)$ space.

\textbf{\textsc{WETW} scoring depends on embedding size.}
Given precomputed embeddings, one \textsc{WETW} comparison costs $\mathcal{O}(dh)$ for a generator of hidden width $h$, including the pair features and the weighted distance. With $h=128$ fixed, this is $\mathcal{O}(d)$ per pair, and scoring the whole database costs $\mathcal{O}(Ndh)$, independent of the series length.

\textbf{\textsc{WETW-FV} limits exact DTW to a fraction of the database.}
\textsc{WETW-FV} computes the $N$ Euclidean distances and sorts them in $\mathcal{O}(N\log N)$ time, rescores the closest $\kappa\rho N$ candidates with \textsc{WETW}, and verifies up to $\rho N$ of them with exact DTW (Section~\ref{sec:wetw_fv}). The worst-case cost of one query is
\[
\mathcal{O}\!\left(T_{\mathrm{enc}}(L)+Nd+\kappa\rho Ndh+N\log N+\rho NL^2\right),
\]
compared with $\mathcal{O}(NL^2)$ for brute-force DTW. With stored embeddings and fixed $d$ and $h$, scoring does not depend on the series length. Because \textsc{WETW} ranks DTW neighbours better than the other embedding-based filters, a smaller shortlist reaches the same recall, which reduces the exact verification work. When this saving exceeds the extra cost of the generator, \textsc{WETW-FV} reaches the target recall faster. Section~\ref{sec:exact_and_ds} measures this trade-off end to end.

\subsection{Training with DTW Regression and Ranking Supervision}
\label{sec:light_heads}

\textbf{Relative distance regression preserves DTW.}
Our DTW-preservation objective $\mathcal{L}_{\mathrm{WD}}$ trains both the encoder and the generator to reproduce DTW distances. Using the length-normalized DTW target $d_{ij}$ defined above and its prediction $\hat d_{ij}$, joint training minimizes the objective
{\small
\begin{equation}
\label{eq:warpdistill}
\mathcal{L}_{\mathrm{WD}}=
\frac{1}{\beta|\mathcal{P}|}\sum_{(i,j)\in\mathcal{P}}
\mathrm{Huber}(r_{ij};\beta),
\end{equation}
}
where $\mathcal{P}$ is the set of supervised pairs and $r_{ij}$ is the signed relative error,
{\small
\[
r_{ij}=\frac{\hat d_{ij}-d_{ij}}{d_{ij}+\varepsilon},
\]
}
with $\varepsilon=10^{-6}$ stabilizing the denominator. The Huber penalty~\cite{huber1964}, with threshold $\beta=0.1$, is defined as
{\small
\[
\mathrm{Huber}(r;\beta)=
\begin{cases}
\frac{1}{2}r^2, & |r|\leq\beta,\\[2pt]
\beta\left(|r|-\frac{\beta}{2}\right), & |r|>\beta.
\end{cases}
\]
}
We use this penalty because its quadratic region encourages precise fitting of small relative errors, while its linear tails limit the influence of large errors, giving a few poorly approximated pairs less influence on training than a purely squared-error loss would.
Relative errors keep distant pairs from dominating and give close pairs more weight than absolute errors would. Following the scaling rationale of~\cite{SEAnet}, for training we divide the square-root DTW cost and predicted embedding distance by $\sqrt{L}$ and $\sqrt{d}$, respectively, to stabilize training. The factor $1/\sqrt{d}$ is absorbed into $s$ in our distance formula.

\textbf{Ranking supervision.}
For the ranking-aware refinement on stored embeddings, we minimize
\begin{equation}
\label{eq:refinement_objective}
\mathcal{L}=\beta\mathcal{L}_{\mathrm{WD}}+\mathcal{L}_{\mathrm{rank}},
\end{equation}
where the factor $\beta$ cancels the $1/\beta$ in~\eqref{eq:warpdistill}, so that the regression term is the plain mean Huber loss. $\mathcal{L}_{\mathrm{WD}}$ fits distance values, but search depends on the order of the nearest neighbours, which relative regression does not enforce. Moreover, most randomly sampled pairs are far apart, so we add a ranking term that focuses on separating true neighbours from other candidates. For an anchor $a$ we sample a true DTW top-10 neighbour $p$ and a negative $n$, which is a series ranked 11 to 2{,}000 by DTW half of the time and a random series otherwise, and use
\[
\mathcal{L}_{\mathrm{rank}}=\max\!\big(0,\ \gamma+\log\hat d_{ap}-\log\hat d_{an}\big),
\qquad \gamma{=}0.1,
\]
which asks the neighbour to be closer than the negative by a relative margin. Since the term compares log-distances, it does not change when all predictions are rescaled. It therefore cannot be satisfied by shrinking every distance, and it leaves the overall scale to the regression term.

\subsection{Optimization Schedule and DTW Supervision}
\label{sec:loss_reg}

\textbf{Sampling reduces DTW supervision costs.}
For a training split of $N_{\mathrm{tr}}$ series, computing DTW for every distinct pair requires $N_{\mathrm{tr}}(N_{\mathrm{tr}}-1)/2$ evaluations, giving $\mathcal{O}(N_{\mathrm{tr}}^2)$ targets and storage. With series length $L$ and fractional warping window radius $w$, the worst-case computation costs $\mathcal{O}(N_{\mathrm{tr}}^2 wL^2)$. We follow the $\mathcal{O}(N_{\mathrm{tr}}\log N_{\mathrm{tr}})$ pair-sampling strategy in~\cite{Lei2019} and adapt this budget to mini-batches by sampling $b_s\lceil\log_2 b_s\rceil$ within-batch pairs for a mini-batch of size $b_s$ and caching their DTW targets per split. For $M$ distinct sampled pairs, target computation and storage scale as $\mathcal{O}(MwL^2)$ and $\mathcal{O}(M)$, respectively, without building a full pairwise matrix. Stage~(iii) ranking also needs each anchor's DTW distances to all training series to find its true neighbours, adding $\mathcal{O}(|\mathcal{A}|N_{\mathrm{tr}}wL^2)$ for an anchor set $\mathcal{A}$ (Table~\ref{tab:offline_cost}).

\textbf{Three training stages.}
Training proceeds in three stages. Stage~(i) warms up the encoder with the MSE between Euclidean embedding distances and DTW targets, with $\omega_\psi$ frozen. This gives a useful distance geometry and stabilizes the embeddings before pair-specific weights are introduced. Stage~(ii) trains $\phi_\theta$ and $\omega_\psi$ jointly with $\mathcal{L}_{\mathrm{WD}}$ alone, so that both adapt to estimating DTW distances. Stage~(iii) introduces ranking supervision in two steps: it first refines the encoder for up to 3K updates, keeping the checkpoint with the best validation recall, with the Euclidean distance and the objective of equation~\eqref{eq:refinement_objective}, so that the stored embeddings themselves order close neighbours better; then, it freezes the encoder and refines the generator on these stored embeddings with the same objective and a cosine learning rate. Finally, for distance fidelity evaluation, we correct systematic scale bias using a scalar fitted on validation pairs and applied unchanged at test time, $\hat d'_{ij}=\hat d_{ij}/(1-b)$ with $b=-\mathbb{E}_{\mathrm{val}}[(\hat d_{ij}-d_{ij})/d_{ij}]$. This centers the signed relative error on the validation pairs. For a fair comparison, the learned baselines are also calibrated using validation pairs, with the fitted correction applied unchanged at test time. We report mean \emph{absolute} relative error to avoid cancellation between signed errors.

\section{\textsc{Experimental Evaluation}}
\label{sec:experiments}

\subsection{Experimental Setup}
\label{sec:exp_setup}

\textbf{Hardware.}
Models are trained on NVIDIA A100 GPUs. Query times are measured on Jean Zay nodes with AMD EPYC 7543 CPUs. For \textsc{WETW-FV}, they include query encoding, Euclidean prefiltering, generator scoring, shortlist selection and exact verification, all on a single CPU thread. The baselines follow the same single-thread protocol, except for NeuralWarp scoring, which is too slow on CPU for our workloads and therefore runs on an A100 GPU.

\textbf{Baselines and implementations.}
\label{sec:baselines}
We compare with LDPS~\cite{ldps}, CDPS~\cite{cdps}, NeuTS~\cite{neuts}, RF-DTAN~\cite{rfdtan}, and NeuralWarp~\cite{grabocka2018neuralwarp} at embedding dimension $d=128$, where the subscript \WD\ marks a method trained with $\mathcal{L}_{\mathrm{WD}}$ instead of its native objective. Each baseline keeps its published architecture and training procedure, and NeuTS uses its published code and configuration. The figures use the $\mathcal{L}_{\mathrm{WD}}$ version of RF-DTAN, whose native objective aligns series to a common average, because it clearly raises validation recall. NeuTS, CDPS and LDPS keep their native objectives, which already approximate DTW, and $\mathcal{L}_{\mathrm{WD}}$ barely changes their recall. For NeuralWarp we use the authors' code~\cite{grabocka2018neuralwarp}, and train the native model with its label-supervised log-loss on the labelled datasets and NeuralWarp$_{\WD}$ with DTW targets. Full-length NeuralWarp comparisons exceed GPU memory on the longest datasets, so we shorten them to at most 625 points with piecewise aggregate approximation (PAA), which averages consecutive blocks of samples, and keep the full series for DTW targets and verification.

\textbf{Exact search and downsampling.}
We implement the UCR-Suite cascade~\cite{rakthanmanon2012searching} in C, combining LB\textsubscript{Kim}, query- and candidate-envelope LB\textsubscript{Keogh}, and early-abandoning DTW within the warping window. We also run LB\textsubscript{Keogh} and LB\textsubscript{Webb}~\cite{webb2021tight} searches, initializing the best-so-far distance to infinity and computing DTW only when a candidate's bound beats it, and MESSI-DTW from DaiSy~\cite{delgaudio2026daisy}, on one core like every other method. For training-free filtering, PAA replaces each block of $g\in\{2,4,8,16,32\}$ consecutive samples with its mean, and DTW ranks the shortened series using the same fractional warping window. We keep the combinations of factor and shortlist size on the Pareto front of recall and time. Averages across datasets use factors $4,8,16$, which were evaluated on every dataset. In scalability experiments, we choose one factor based on validation recall and a short timing run. In all cases, the UCR-Suite cascade verifies shortlisted candidates using the original, full-resolution series.

\textbf{\textsc{WESee} and \textsc{WETW} hyperparameters.}
The encoder stacks six residual blocks with 64 channels and kernel size 7, doubling the dilation from one block to the next, and ends with an MLP that projects to a 128-dimensional embedding. The TFE applies temporal heads with kernel sizes 3, 45 and 95 and pools their output to 256 steps. The generator is an MLP with a hidden width of 128. Stage~(i) trains the encoder for 50 epochs on up to 5K random pairs per epoch. Stage~(ii) trains encoder and generator jointly for 100 epochs on $b_s\lceil\log_2 b_s\rceil$ within-minibatch pairs per minibatch of size $b_s$, after which the generator continues alone until the validation loss stops improving. Encoder and generator use AdamW ($\mathrm{lr}=10^{-4}$ and $5\times10^{-5}$) with batches of 256. Stage~(iii) refines the encoder for 3K updates at $\mathrm{lr}=10^{-4}$, then the generator for 30K or 100K updates at $\mathrm{lr}=10^{-3}$ with a cosine schedule, each update using 4{,}096 regression pairs and 1{,}024 ranking triplets, and keeps the checkpoints with the best validation recall. Over three training seeds, \textsc{WETW-FV} at a 2\% shortlist reaches a mean recall@1 of $0.948\pm0.004$ and recall@10 of $0.896\pm0.008$.

\textbf{Selecting the prefilter ratio $\kappa$.}
For the multi-dataset search experiments, we select the smallest $\kappa\in\{1,2,3,5,10,20\}$, or rescoring of the whole database, that keeps both validation recall@1 and recall@10 within $0.005$ of the best value achieved for each measure over the tested settings. This limits rescoring work while retaining the best recall. If no setting meets both thresholds, we choose the one with the highest sum of recall@1 and recall@10. At a 2\% shortlist, every dataset selects rescoring, with $\kappa$ between 5 and 20, and none selects full-database rescoring.

\textbf{Evaluation measures.}
Let $\mathcal{Q}$ be the set of queries, $G_k(q)$ be the exact DTW top-$k$ neighbours of query $q$, and $S_\rho(q)$ its shortlist at fraction $\rho$. We measure neighbour preservation by
{\small
\[
\operatorname{Recall}@k
=\frac{1}{|\mathcal{Q}|}\sum_{q\in\mathcal{Q}}
\frac{|G_k(q)\cap S_\rho(q)|}{k},
\]
}
reporting $k=1$ and $k=10$ and comparing methods at equal shortlist fractions. The $\pm$ in tables is the standard error $\sqrt{p(1-p)/100}$ of a recall@1 $p$, and $\sqrt{\sum_i \mathrm{se}_i^2}/8$ for averages. For held-out pairs $\mathcal{P}_{\mathrm{test}}$ with DTW targets $d_{ij}$, distance fidelity is the mean absolute relative error
{\small
\[
\operatorname{Mean}(|\mathrm{RE}|)
=\frac{1}{|\mathcal{P}_{\mathrm{test}}|}
\sum_{(i,j)\in\mathcal{P}_{\mathrm{test}}}
\frac{|\hat d'_{ij}-d_{ij}|}{d_{ij}}.
\]
}
We abbreviate this mean as $|\mathrm{RE}|$ in the rest of the paper. For a query $q$, let $t(q)$ be the time in seconds from receiving $q$ to returning its neighbours, including encoding, database scoring, shortlist selection, and exact verification. Over a set of timed queries $\mathcal{Q}_{\mathrm{time}}$, the mean time per query is
{\small
\[
\bar t=\frac{1}{|\mathcal{Q}_{\mathrm{time}}|}
\sum_{q\in\mathcal{Q}_{\mathrm{time}}}t(q).
\]
}

\textbf{Datasets.}
\label{sec:datasets}
Table~\ref{tab:datasets} summarizes the whole-series search datasets, with series lengths from 256 to 48K and database sizes from 20.9K to 840K series. Each dataset has 100 held-out search queries, with disjoint training, validation, test, database, and query splits. The Monash MONSTER archive~\cite{dempster2025monster} datasets (WhaleSounds, CornellWhale, MosquitoSound, FruitFlies, AudioMNIST-DS~\cite{becker2024audiomnist}, DREAMER-A~\cite{katsigiannis2018dreamer}, and LenDB~\cite{magrini2020lendb}) provide class labels, whereas Seismic is unlabelled. Native NeuralWarp is evaluated on the seven labelled datasets. Training and validation support model fitting and selection, the test split held-out distance evaluation, and the queries retrieval against the database. Ground-truth DTW distances use \texttt{dtaidistance}~\cite{meert2022dtaidistance} with \texttt{window}$=\lfloor 0.1L\rfloor$.

\begin{table}[t]
\centering
\caption{Dataset splits (Train/Val./Test), database size (DB), search queries (Q), and series length ($L$).}
\label{tab:datasets}
\footnotesize
\setlength{\tabcolsep}{2pt}
\renewcommand{\arraystretch}{1.12}
\begin{tabular*}{\columnwidth}{@{\extracolsep{\fill}}lrrrrrr@{}}
\toprule
\textbf{Dataset} & \textbf{Train} & \textbf{Val.} & \textbf{Test} & \textbf{DB} & \textbf{Q} & $L$ \\
\midrule
Seismic~\cite{iris2018seismic} & 100K & 50K & 10K & 840K & 100 & 256 \\
WhaleSounds~\cite{dempster2025monster} & 30K & 10K & 10K & 40K & 100 & 2.5K \\
CornellWhale~\cite{dempster2025monster} & 20K & 4.5K & 4.5K & 24.5K & 100 & 4K \\
MosquitoSound~\cite{dempster2025monster} & 50K & 10K & 10K & 60K & 100 & 3.75K \\
FruitFlies~\cite{dempster2025monster} & 20K & 6.8K & 6.8K & 26.8K & 100 & 5K \\
DREAMER-A~\cite{dempster2025monster} & 20K & 5K & 5K & 60K & 100 & 256 \\
LenDB~\cite{dempster2025monster} & 20K & 5K & 5K & 60K & 100 & 540 \\
AudioMNIST-DS~\cite{dempster2025monster} & 5K & 2K & 2K & 20.9K & 100 & 4K \\
AudioMNIST~\cite{dempster2025monster} & 2K & 300 & 300 & 26K & 100 & 48K \\
\bottomrule
\end{tabular*}
\end{table}

\input{res/rev2/experimental_results}

\section{Conclusions}
\label{sec:conclusions}
\textbf{Pair-specific comparisons make learned DTW filtering effective.}
We introduced \textsc{WETW}, a pair-specific comparison over series embeddings, with the \textsc{WESee} encoder and ranking-aware DTW supervision. Three findings stand out: (i)~pair-specific weights give the most accurate DTW approximation on every dataset, halving the error of the best static measure ($0.04$ against $0.08$) while keeping the measure non-negative, symmetric and zero on identical embeddings, unlike a direct MLP on the same features; (ii)~embeddings must preserve when patterns occur, as at matched capacity the TFE raises recall@1 from $0.90$ to $0.95$; and (iii)~fitting DTW distances is not enough, as regression alone does not necessarily preserve the order of the closest neighbours, and ranking regularization raises recall@1 from $0.79$ to $0.95$. 

Together, these components make search both fast and accurate, as \textsc{WETW-FV} verifies a small percentage of the database on one CPU thread and reaches a mean recall@10 of $0.93$ at $11.4\times$ the speed of the fastest exact search, against $0.59$ for the strongest learned baseline. On 26K series of 48K points, \textsc{WETW-FV} remains $40\times$ faster than the UCR Suite at a recall@10 of $0.91$, against at most $0.60$ for the learned baselines, and small training sets keep its supervision cost low.

\textbf{Lower costs, richer series and indexed search remain open.}
We leave three extensions to future work: (i) cutting the offline supervision and scoring costs through informative training pairs and more compact generators, (ii) widening the method to multivariate and variable-length series, 
and (iii) transferring trained models across datasets and under distribution shift.

\clearpage
\section*{Acknowledgments and Disclosure of Funding}
Supported by the EU Horizon projects TwinODIS (101160009) and DataGEMS (101188416). 
This work was granted access to the HPC resources of IDRIS under the allocations 2025-A0191012641, 2025-AD011012641R4, 2025-AD011016632, and 2026-AD011016632R1 by GENCI.
We thank Azedine Mani, Head of the IT team (UFR Math\'ematiques et Informatique, Universit\'e Paris Cit\'e), for his pivotal role in creating and making available computational resources essential to this work.
The authors declare no competing interests.

\textbf{Use of AI-generated content.} The authors used the AI coding assistants Claude Code (Anthropic, Claude Opus 5.5) and Codex (OpenAI, GPT-6.1 Sol), which assisted with implementing the experimental evaluation code and the scripts for the tables and figures in Section~\ref{sec:experiments}, and with preparing the text. The authors designed the method and the experiments, reviewed the generated content, verified the reported results, and take full responsibility for this paper.

\bibliographystyle{IEEEtran}
\bibliography{references}

\FloatBarrier
\end{document}

%% file: res/rev2/experimental_results.tex
\subsection{Results and Discussion}
\label{sec:main_results}

\label{sec:retrieval_results}
\label{sec:repro}
\label{sec:distance_fidelity}
\label{sec:encoders}
\label{sec:ablations}
\label{sec:alignment_probe}
\label{sec:diagnostic}
\label{sec:length_scaling}
\label{sec:wetw_search_efficiency}
\label{sec:exact_and_ds}
\label{sec:generator_cost}
\label{sec:breakeven}
\label{sec:regime_check}

\begin{figure*}[t]
\centering
\includegraphics{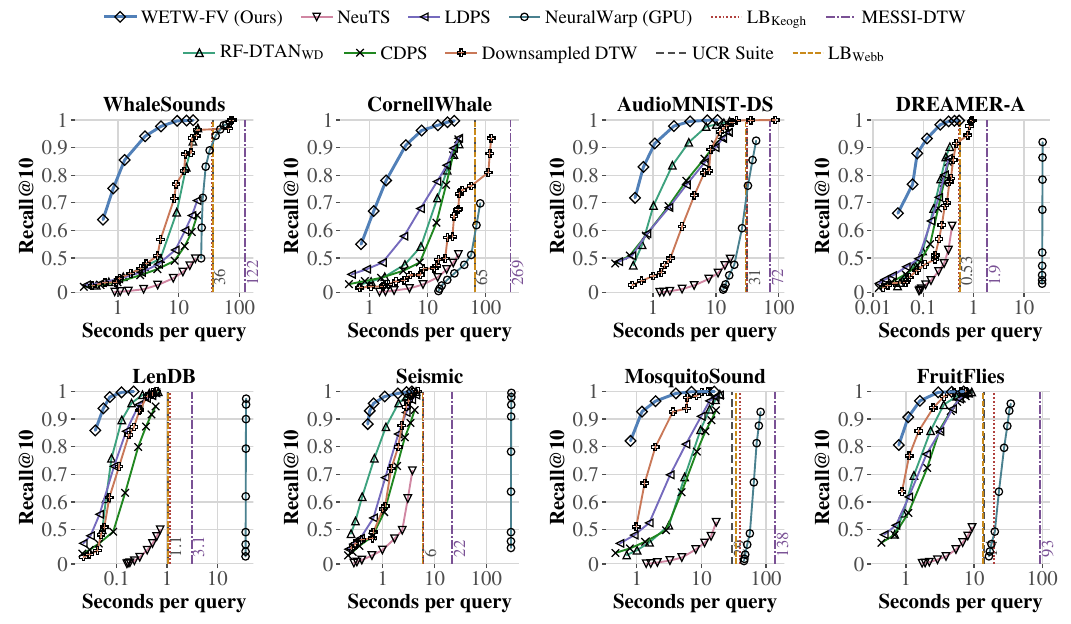}
\caption{End-to-end search, recall@10 against seconds per query. CPU stages use one thread; NeuralWarp scoring uses an A100 GPU. Vertical lines mark exact searches.}
\label{fig:e2e_all}
\end{figure*}

\textbf{\textsc{WETW-FV} is fastest where lower bounds prune little.}
Which filter gives the fastest search at a given recall? For each query, every filter scores the whole database, keeps a shortlist of nine sizes from 0.5\% to 50\% of the database, and verifies it with the same exact cascade, the UCR-Suite lower bounds with EAPrunedDTW~\cite{herrmann2021eap}. The UCR Suite exact search uses the same cascade over the whole database, which is $2.4\times$ to $3.7\times$ faster than with the original UCR-Suite DTW~\cite{rakthanmanon2012searching}. Recall@10 is measured over the 100 queries, and time counts query encoding, database scoring, shortlist selection and verification. Figure~\ref{fig:e2e_all} reports all eight datasets (their average appears in Figure~\ref{fig:main}). Speed-ups are measured against the fastest exact method on each dataset, as no exact method dominates on every dataset. Searching with a single lower bound, LB\textsubscript{Keogh} or LB\textsubscript{Webb}, takes about as long as the full UCR Suite cascade (from $1.05\times$ faster to $1.4\times$ slower), while MESSI-DTW is $2.4\times$ to $6.9\times$ slower. \textsc{WETW-FV} reaches a recall@10 of $0.9$ within a 10\% shortlist on all eight datasets. At this recall it is faster than exact search by $12.8\times$ on WhaleSounds, $15.3\times$ on CornellWhale, $27.7\times$ on AudioMNIST-DS, $4.3\times$ on DREAMER-A, $19.3\times$ on LenDB, $10.4\times$ on Seismic, $24.5\times$ on MosquitoSound and $12.4\times$ on FruitFlies. It is also the fastest filter at this recall on every dataset, $2.3\times$ to $8.0\times$ faster than the second best, which differs between datasets. NeuTS does not reach this recall on any dataset, and NeuralWarp, where it does, is no faster than exact search, even though it ranks on a GPU. Averaged over the datasets, \textsc{WETW-FV} reaches a mean recall@10 of $0.93$ at $11.4\times$ the speed of the fastest exact search, while at the same mean query time the baselines reach at most $0.59$. The gain is smallest on DREAMER-A, whose short series ($L=256$) make exact search fast ($0.5$\,s per query) and leave ranking with most of the query time. Overall, \textsc{WETW-FV} gives the fastest end-to-end search at high recall on all eight datasets.

\begin{figure*}[t]
\centering
\includegraphics[width=0.85\textwidth]{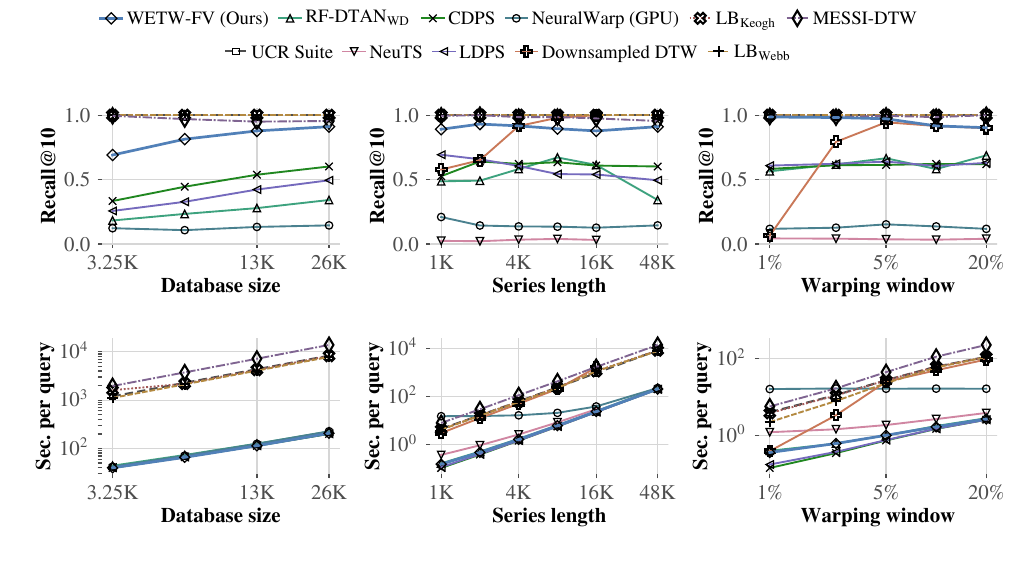}
\vspace*{-0.4cm}
\caption{Scalability on AudioMNIST. Recall@10 and seconds per query at a 2\% shortlist against database size, series length and warping window. Fixed settings: (left) $L{=}48$K and window $10\%$; (middle) $N{=}26$K and window $10\%$; (right) $N{=}26$K and $L{=}4$K. Windows are fractions of series length.}
\label{fig:amnist}
\end{figure*}

\textbf{Lower bounds stay loose where series oscillate within the warping window.}
Why do lower bounds help so little on some datasets? We run the exact search with each bound and record the share of candidates that reach the DTW stage, where early abandoning may still stop the computation (Table~\ref{tab:lower_bound_survival}). With LB\textsubscript{Keogh}, more than 90\% of candidates reach DTW on five of the eight datasets. The tighter LB\textsubscript{Webb} removes about a fifth more candidates on four datasets, yet about half or more still reach DTW everywhere except FruitFlies. This extra pruning also saves little time, likely because early-abandoning DTW would soon stop the same candidates anyway, and only on FruitFlies does LB\textsubscript{Webb} cut query time by about a third. Both bounds compare the query with the candidate's envelope over the warping window, and query points inside that envelope add nothing to the bound. When series oscillate within the window, the envelope covers most of the signal's range and nearly every query point falls inside it. This is the case on WhaleSounds, CornellWhale and DREAMER-A, where LB\textsubscript{Webb} gains almost nothing, while fewer points fall inside on the datasets where it helps. Overall, on oscillating series lower bounds leave most candidates to the DTW stage, which is the work a learned filter avoids.

\input{res/rev2/table_lower_bound_survival}

\textbf{\textsc{WETW-FV} scales to larger databases, longer series, and wider warping windows.}
Does \textsc{WETW-FV} remain fast and accurate as the search problem grows? We study scalability on AudioMNIST with 2K training series, varying database size from 3.25K to 26K, series length from 1K to 48K and warping window from $1\%$ to $20\%$ of the length, one factor at a time. Figure~\ref{fig:amnist} reports recall@10 and query time at a $2\%$ shortlist. Every learned method except NeuralWarp maps each series to a 128-dimensional embedding, compressing 48K values in the left panels, up to 48K in the middle panels and 4K in the right panels. The exact searches return the exact neighbours, but \textsc{WETW-FV} is on average $31\times$ to $33\times$ faster than the UCR Suite and the single lower bounds and $56\times$ faster than MESSI-DTW. Learned baselines run in similar or slightly shorter time but generally reach lower recall, and NeuralWarp stays much slower even though it ranks on an A100 GPU. The largest setting, 26K series of length 48K, holds about 1.2 billion values (six times as many as Seismic). 
In this setting, \textsc{WETW-FV} reaches a recall@10 of $0.91$ against at most $0.60$ for the learned baselines, and a query takes about $3.4$ minutes, still $40\times$ faster than the UCR Suite. NeuTS is not applicable at this length, since its recurrent encoder processes each series one time step at a time, so one training iteration over series of 48K points takes 192\,s, which makes training infeasible. It is therefore absent from the database-size panels and at 48K points. As \textsc{WETW-FV} runs DTW only on the $\rho N$ shortlisted series (Section~\ref{sec:complexity}), it can be up to $1/\rho=50\times$ faster at a 2\% shortlist when lower bounds prune little, close to the measured $40\times$ at 48K points.

\begin{figure}[t]
\centering
\includegraphics{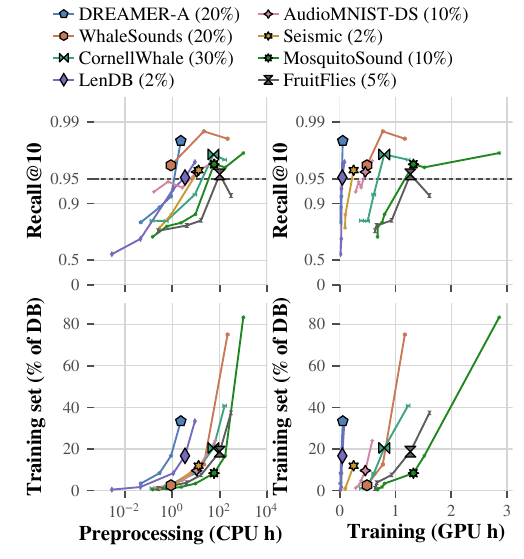}
\caption{Recall@10 and training-set size versus preprocessing and training costs. Highlighted markers mark the smallest set reaching $0.95$ recall, at the shortlist fraction in the legend.}
\label{fig:trainsize}
\label{fig:trainsize_cost}
\end{figure}

\textbf{A few thousand series are enough to train \textsc{WETW} on long series.}
How much training data does \textsc{WETW} need, and what does it cost? We retrain \textsc{WESee} and \textsc{WETW} on subsets of 250 series up to the full training set, keeping the validation set, database and query set fixed. For each dataset, we fix the shortlist to the smallest fraction at which \textsc{WETW} reaches recall@10 of 0.95 (see Figure~\ref{fig:trainsize}). We then find the smallest tested training subset that reaches recall@10 of 0.95 at that fraction. Preprocessing counts the DTW pairs used for training, including distances needed for ranking regularization. On average, a training set of about 15\% of the database reaches recall@10 of 0.95. At these sizes, training takes less than 1.5 GPU hours on every dataset. Most datasets come close to their full-data recall with 5K to 10K training series, while DREAMER-A and Seismic keep improving up to their full training sets. The required training fraction therefore varies, from 2.5\% of the database on WhaleSounds to 33.3\% on DREAMER-A. Table~\ref{tab:offline_cost} reports, at these training sizes, the CPU hours to generate the DTW targets that training reads and the GPU hours to fit the models. Saved s/q is the search time saved per query, $\Delta t=\bar t_{\mathrm{exact}}-\bar t_{\textsc{WETW-FV}}$, where $\bar t_{\mathrm{exact}}$ is the time of the fastest exact search and $\bar t_{\textsc{WETW-FV}}$ that of \textsc{WETW-FV} at its fastest shortlist with recall@10 of at least 0.9, both on one CPU thread. For a target generation time $T_{\mathrm{CPU}}$, the break-even point $\mathrm{BE}=T_{\mathrm{CPU}}/\Delta t$ is the number of queries after which the time saved exceeds the one-time cost of the targets. These targets are paid once, by \textsc{WETW} as by every learned baseline, while exact search repeats its DTW work for every query. The target cost is repaid after 95 queries on WhaleSounds and at most 29.2K on FruitFlies, with a median of about 8K. Beyond that point every query adds its saving, so each further 1K queries saves $5.9$ hours of search time on average over the eight datasets, from $6.5$ minutes on DREAMER-A, whose short series make exact search fast, to $16.8$ hours on CornellWhale.
\input{res/rev2/table_offline_cost}

\begingroup\clubpenalties 3 10000 10000 0 
\textbf{Pair-specific weights give the most accurate distances.}
Does \textsc{WETW} need pair-specific weights, or would a trained static distance on the same embeddings suffice? We freeze each dataset's \textsc{WESee} embeddings and compare \textsc{WETW} with three static distances, all trained with the same regression objective and budget and without ranking regularization (Table~\ref{tab:heads}), so that every measure receives the same supervision, which is why \textsc{WETW} reaches a recall@1 of $0.79$ here, against $0.95$ with ranking regularization (Table~\ref{tab:components}). Euclidean distance uses the embeddings as they are and shows what the encoder alone captures. The diagonal measure learns one weight per dimension shared by all pairs (the same form as \textsc{WETW} without pair specificity), and the full-matrix measure $\|P(\mathbf e_i-\mathbf e_j)\|_2$ can also rotate and mix dimensions, making it the most flexible static distance. \textsc{WETW} has the lowest error on every dataset ($0.04$ on average, against $0.08$ for the full matrix and $0.24$ for Euclidean distance) and the highest mean recall@1 ($0.79$, against at most $0.73$), with the best or tied-best recall@1 on six of the eight datasets. Its gap to the diagonal measure ($0.04$ against $0.13$ in error, $0.79$ against $0.72$ in recall@1) isolates the benefit of adapting the weights to each pair, and is largest on WhaleSounds, where \textsc{WETW} retrieves the nearest neighbour for $70\%$ of queries against at most $28\%$. On AudioMNIST-DS and FruitFlies a static measure retrieves it slightly more often ($0.89$ against $0.82$ and $0.86$ against $0.83$), as the lower error of \textsc{WETW} there comes from distant pairs while it overestimates the distances to the closest candidates. Pair-specific weights are therefore needed for both accurate distances and accurate search.\par\endgroup

\input{res/rev2/table_heads}

\textbf{The weighted form preserves measure properties.}
Would a network that predicts the distance directly from the same input features work as well? We train an MLP of the same width on the same symmetric features $[|\mathbf e_i-\mathbf e_j|,\mathbf e_i\odot\mathbf e_j,\mathbf e_i+\mathbf e_j]$ to output the distance itself, initialized at the Euclidean distance and trained with the same budget and objectives as \textsc{WETW}, and use it in place of \textsc{WETW} in \textsc{WETW-FV} (Table~\ref{tab:direct_mlp}). \textsc{WETW} reaches a slightly higher mean recall@1 ($0.95$ against $0.93$) and matches or beats the MLP on every dataset, most clearly on CornellWhale ($0.91$ against $0.85$). It also has a better structure, as it only reweights Euclidean coordinate differences with positive weights, so its distance is non-negative, symmetric and zero exactly when two embeddings coincide. The direct MLP is not tied to these properties. Its distance from a series to itself, $\hat d_{ii}$, is on average $0.63$ times the DTW distance $d_{\mathrm{NN}}$ from that series to its nearest neighbour, against exactly zero for \textsc{WETW}. Recall does not reveal this, since the queries are not part of the database and recall@k ignores the order within the shortlist. As the weighted form keeps these properties at no extra cost, we keep \textsc{WETW}.

\input{res/rev2/table_direct_mlp}

\textbf{Fitting DTW distances is not enough, ranking regularization is essential.}
Is fitting DTW distances enough to preserve the order of neighbours, which is what search uses? We train every learned encoder with $\mathcal{L}_{\mathrm{WD}}$ and compare its embeddings in two ways, with Euclidean distance and with \textsc{WETW}. Table~\ref{tab:components} reports both distances after refinement with the regression term alone ($\mathcal{L}_{\mathrm{WD}}$) and with ranking regularization added ($\mathcal{L}$). Ranking regularization first fine-tunes the encoder for 3K updates, keeping the checkpoint with the best validation recall. Both distances then use this ranking-refined encoder, Euclidean distance directly and \textsc{WETW} after its generator is refined on the frozen encoder. Training on DTW targets matters mainly for encoders not designed for DTW, as $\mathcal{L}_{\mathrm{WD}}$ raises recall@1 of RF-DTAN with Euclidean distance from $0.36$ to $0.52$ and of NeuralWarp with its own network from $0.06$ to $0.31$, and leaves NeuTS, CDPS and LDPS, which already approximate DTW, nearly unchanged. Regression alone fits distances well but orders neighbours poorly. On \textsc{WESee}, regression-only \textsc{WETW} reaches a mean relative error of $0.04$ yet recall@1 of only $0.79$, because close competitors lie within a few percent of the true neighbour and the mean error mostly reflects the many distant pairs. Adding ranking regularization raises the recall@1 of \textsc{WETW} for every encoder, to $0.95$ on \textsc{WESee} and by $0.16$ to $0.21$ on RF-DTAN, CDPS and LDPS, while its error grows by at most $0.02$, and NeuralWarp improves in the same way from $0.31$ to $0.51$ when its own warping network is refined. Accurate distances are therefore not sufficient for search, and ranking supervision is needed on top of them.

\textbf{Pair-specific weights and ranking supervision are complementary.}
Does \textsc{WETW} add to ranking regularization, or does ranking alone explain the gain? Under regression alone, \textsc{WETW} raises recall@1 over Euclidean distance for \textsc{WESee} and RF-DTAN, and on NeuTS, CDPS and LDPS it roughly halves the error but changes recall by at most $0.02$. Ranking applied to Euclidean distance helps \textsc{WESee} and RF-DTAN but leaves CDPS and LDPS unchanged, while ranking applied to \textsc{WETW} improves all of them. With ranking regularization, \textsc{WETW} is the best distance for every encoder, ahead of ranking-trained Euclidean distance by $0.10$ on \textsc{WESee} and by $0.16$ to $0.24$ on RF-DTAN, CDPS and LDPS. Neither component alone reaches the full gain, as pair-specific weights provide the flexibility to reorder close neighbours and ranking regularization trains the generator to use it. \textsc{WESee} benefits most, suggesting that its global summary and multi-scale temporal features give the generator more to work with.

\input{res/rev2/table_components}

\textbf{The TFE raises recall on every dataset.}
Does the TFE add information that the pooled ResNet summary misses? Max pooling keeps which patterns a series contains but discards when they occur, and this timing is what DTW aligns. We train the encoder with and without the TFE, widening the no-TFE backbone to match the with-TFE backbone within $1\%$ in encoder parameters. Both output 128 dimensions and receive the same training and evaluation (Table~\ref{tab:architecture_capacity_matched}). The TFE raises recall@1 and lowers or matches the error on every dataset, raising mean recall@1 from $0.90$ to $0.95$. The largest gains are on CornellWhale ($+0.16$) and WhaleSounds ($+0.09$), followed by MosquitoSound ($+0.07$) and FruitFlies ($+0.04$), on both of which it also lowers the error from $0.10$ to $0.06$, while DREAMER-A, AudioMNIST-DS, LenDB and Seismic gain $0.01$ to $0.02$. Recall@10 follows the same pattern except on AudioMNIST-DS, where the encoder without the TFE is ahead ($0.95$ against $0.92$). Since a wider pooled backbone with the same number of parameters does not recover this gain, it comes from the temporal information that pooling discards rather than from extra capacity, so the TFE improves search by preserving when patterns occur, most of all on long series.

\input{res/rev2/table_architecture_capacity_matched}

%% file: res/rev2/table_lower_bound_survival.tex
\begin{table}[t]
\centering
\caption{Candidates reaching the DTW stage (\%) with each lower bound used alone. Lower is better.}
\label{tab:lower_bound_survival}
\small
\setlength{\tabcolsep}{12pt}
\begin{tabular}{@{}lrr@{}}
\toprule
Dataset & LB\textsubscript{Keogh} & LB\textsubscript{Webb} \\
\midrule
WhaleSounds & 99 & 98 \\
CornellWhale & 95 & 94 \\
AudioMNIST-DS & 93 & 87 \\
DREAMER-A & 93 & 90 \\
LenDB & 69 & 51 \\
Seismic & 94 & 72 \\
MosquitoSound & 66 & 46 \\
FruitFlies & 45 & 24 \\
\bottomrule
\end{tabular}
\end{table}

%% file: res/rev2/table_offline_cost.tex
\begin{table}[t]
\centering
\caption{Saved s/q: \textsc{WETW-FV} time saved/query vs. exact search at recall@10 $\geq0.9$. BE: number of queries to break even.}
\label{tab:offline_cost}
\footnotesize
\setlength{\tabcolsep}{2pt}
\begin{tabular*}{\columnwidth}{@{\extracolsep{\fill}}lrrrrr@{}}
\toprule
Dataset & $N_{\mathrm{tr}}$ & CPU h & GPU h & Saved s/q & BE \\
\midrule
WhaleSounds & 1K & 0.9 & 0.5 & 33.1 & 95 \\
CornellWhale & 5K & 54 & 0.8 & 60.4 & 3.2K \\
AudioMNIST-DS & 2K & 11 & 0.5 & 28.5 & 1.3K \\
DREAMER-A & 20K & 2.3 & 0.06 & 0.4 & 21.0K \\
LenDB & 10K & 3.4 & 0.05 & 1.0 & 12.5K \\
Seismic & 100K & 13 & 0.2 & 5.3 & 8.7K \\
MosquitoSound & 5K & 58 & 1.3 & 28.3 & 7.4K \\
FruitFlies & 5K & 100 & 1.3 & 12.3 & 29.2K \\
\midrule
Average & -- & 30 & 0.6 & 21.2 & 10.4K \\
\bottomrule
\end{tabular*}
\end{table}

%% file: res/rev2/table_heads.tex
\begin{table}[t]
\centering
\caption{Distance function ablations on frozen \textsc{WESee} embeddings: mean $|\mathrm{RE}|$ / recall@1 at a 2\% shortlist. Bold marks row-wise best values, including ties.}
\label{tab:heads}
\footnotesize
\setlength{\tabcolsep}{2pt}
\begin{tabular*}{\columnwidth}{@{\extracolsep{\fill}}lcccc@{}}
\toprule
Dataset & Euclidean & Diagonal & Full matrix & \textsc{WETW} \\
\midrule
WhaleSounds & 0.31\,/\,0.18\se{0.04} & 0.21\,/\,0.23 & 0.09\,/\,0.28 & \textbf{0.04}\,/\,\textbf{0.70}\se{0.05} \\
CornellWhale & 0.38\,/\,\textbf{0.37}\se{0.05} & 0.20\,/\,0.36 & 0.09\,/\,0.36 & \textbf{0.03}\,/\,\textbf{0.37}\se{0.05} \\
AudioMNIST-DS & 0.23\,/\,0.85\se{0.04} & 0.12\,/\,\textbf{0.89} & 0.08\,/\,\textbf{0.89} & \textbf{0.06}\,/\,0.82\se{0.04} \\
DREAMER-A & 0.16\,/\,0.71\se{0.05} & 0.08\,/\,0.76 & 0.05\,/\,0.80 & \textbf{0.03}\,/\,\textbf{0.84}\se{0.04} \\
LenDB & 0.11\,/\,0.96\se{0.02} & 0.07\,/\,0.95 & 0.06\,/\,0.94 & \textbf{0.04}\,/\,\textbf{0.98}\se{0.01} \\
Seismic & 0.29\,/\,0.61\se{0.05} & 0.09\,/\,0.91 & 0.06\,/\,0.95 & \textbf{0.02}\,/\,\textbf{0.96}\se{0.02} \\
MosquitoSound & 0.19\,/\,0.82\se{0.04} & 0.12\,/\,0.78 & 0.09\,/\,0.82 & \textbf{0.07}\,/\,\textbf{0.86}\se{0.03} \\
FruitFlies & 0.25\,/\,0.80\se{0.04} & 0.14\,/\,\textbf{0.86} & 0.09\,/\,0.83 & \textbf{0.07}\,/\,0.83\se{0.04} \\
\midrule
Average & 0.24\,/\,0.66\se{0.01} & 0.13\,/\,0.72 & 0.08\,/\,0.73 & \textbf{0.04}\,/\,\textbf{0.79}\se{0.01} \\
\bottomrule
\end{tabular*}
\end{table}

%% file: res/rev2/table_direct_mlp.tex
\begin{table}[t]
\centering
\caption{\textsc{WETW} vs.\ a direct MLP in \textsc{WETW-FV}.}
\label{tab:direct_mlp}
\footnotesize
\setlength{\tabcolsep}{1.5pt}
\begin{tabular*}{\columnwidth}{@{\extracolsep{\fill}}lcccc@{}}
\toprule
 & \multicolumn{2}{c}{R@1} & \multicolumn{2}{c}{$\hat d_{ii}/d_{\mathrm{NN}}$} \\
\cmidrule(lr){2-3}\cmidrule(lr){4-5}
Dataset & \textsc{WETW} & MLP & \textsc{WETW} & MLP \\
\midrule
WhaleSounds & \textbf{0.89}\se{0.03} & \textbf{0.89}\se{0.03} & \textbf{0.00} & 0.58 \\
CornellWhale & \textbf{0.91}\se{0.03} & 0.85\se{0.04} & \textbf{0.00} & 0.49 \\
AudioMNIST-DS & \textbf{0.98}\se{0.01} & 0.96\se{0.02} & \textbf{0.00} & 0.73 \\
DREAMER-A & \textbf{0.85}\se{0.04} & 0.84\se{0.04} & \textbf{0.00} & 0.67 \\
LenDB & \textbf{1.00}\se{0.00} & \textbf{1.00}\se{0.00} & \textbf{0.00} & 0.65 \\
Seismic & \textbf{0.99}\se{0.01} & 0.98\se{0.01} & \textbf{0.00} & 0.52 \\
MosquitoSound & \textbf{0.98}\se{0.01} & 0.94\se{0.02} & \textbf{0.00} & 0.62 \\
FruitFlies & \textbf{0.98}\se{0.01} & 0.96\se{0.02} & \textbf{0.00} & 0.78 \\
\midrule
Average & \textbf{0.95}\se{0.01} & 0.93\se{0.01} & \textbf{0.00} & 0.63 \\
\bottomrule
\end{tabular*}
\end{table}

%% file: res/rev2/table_components.tex
\begin{table}[t]
\centering
\caption{Objective ablations. Cells show mean $|\mathrm{RE}|$ / recall@1 at a 2\% shortlist. Bold marks the best value per encoder group and measure, including ties.}
\label{tab:components}
\footnotesize
\setlength{\tabcolsep}{2pt}
\begin{tabular*}{\columnwidth}{@{\extracolsep{\fill}}llccc@{}}
\toprule
Encoder & Distance & Native & $\mathcal{L}_{\mathrm{WD}}$ & $\mathcal{L}$ \\
\midrule
\textsc{WESee} & Euclidean & -- & 0.24\,/\,0.66\se{0.01} & 0.11\,/\,0.85\se{0.01} \\
 & \textsc{WETW} & -- & \textbf{0.04}\,/\,0.79\se{0.01} & 0.05\,/\,\textbf{0.95}\se{0.01} \\
\midrule
RF-DTAN & Euclidean & 0.57\,/\,0.36\se{0.02} & 0.21\,/\,0.52\se{0.01} & 0.22\,/\,0.58\se{0.01} \\
 & \textsc{WETW} & -- & \textbf{0.07}\,/\,0.61\se{0.02} & 0.08\,/\,\textbf{0.82}\se{0.01} \\
\midrule
NeuTS & Euclidean & 0.31\,/\,0.05\se{0.01} & 0.37\,/\,0.04\se{0.01} & 0.35\,/\,0.07\se{0.01} \\
 & \textsc{WETW} & -- & \textbf{0.18}\,/\,0.06\se{0.01} & 0.19\,/\,\textbf{0.10}\se{0.01} \\
\midrule
CDPS & Euclidean & 0.24\,/\,0.38\se{0.02} & 0.24\,/\,0.36\se{0.02} & 0.24\,/\,0.36\se{0.02} \\
 & \textsc{WETW} & -- & \textbf{0.11}\,/\,0.36\se{0.02} & 0.13\,/\,\textbf{0.52}\se{0.02} \\
\midrule
LDPS & Euclidean & 0.17\,/\,0.49\se{0.02} & 0.16\,/\,0.50\se{0.02} & 0.16\,/\,0.50\se{0.02} \\
 & \textsc{WETW} & -- & \textbf{0.08}\,/\,0.50\se{0.02} & 0.09\,/\,\textbf{0.69}\se{0.02} \\
\midrule
NeuralWarp$^{\dagger}$ & Own network & 0.54\,/\,0.06\se{0.01} & \textbf{0.09}\,/\,0.31\se{0.01} & 0.10\,/\,\textbf{0.51}\se{0.02} \\
\midrule
Average & Euclidean & 0.32\,/\,0.32\se{0.01} & 0.25\,/\,0.42\se{0.01} & 0.22\,/\,0.47\se{0.02} \\
 & \textsc{WETW} & -- & \textbf{0.10}\,/\,0.47\se{0.01} & 0.11\,/\,\textbf{0.62}\se{0.01} \\
\bottomrule
\end{tabular*}
\par\smallskip{\footnotesize\raggedright $^{\dagger}$NeuralWarp uses its own distance and is excluded from averages.\par}
\end{table}

%% file: res/rev2/table_architecture_capacity_matched.tex
\begin{table}[t]
\centering
\caption{Temporal branch ablation. Result cells show $|\mathrm{RE}|$ / recall@1 of \textsc{WETW-FV} at a 2\% shortlist. Bold marks the better value per row and measure, including ties.}
\label{tab:architecture_capacity_matched}
\footnotesize
\setlength{\tabcolsep}{2pt}
\begin{tabular*}{\columnwidth}{@{\extracolsep{\fill}}lrrcc@{}}
\toprule
 & \multicolumn{2}{c}{Params.} & \multicolumn{2}{c}{$|\mathrm{RE}|$ / R@1} \\
\cmidrule(lr){2-3}\cmidrule(lr){4-5}
Dataset & No TFE & With TFE & No TFE & With TFE \\
\midrule
WhaleSounds & 2.95M & 2.95M & \textbf{0.06}\,/\,0.80\se{0.04} & \textbf{0.06}\,/\,\textbf{0.89}\se{0.03} \\
CornellWhale & 2.95M & 2.95M & 0.07\,/\,0.75\se{0.04} & \textbf{0.06}\,/\,\textbf{0.91}\se{0.03} \\
AudioMNIST-DS & 2.95M & 2.95M & 0.07\,/\,0.97\se{0.02} & \textbf{0.06}\,/\,\textbf{0.98}\se{0.01} \\
DREAMER-A & 563K & 558K & \textbf{0.05}\,/\,0.83\se{0.04} & \textbf{0.05}\,/\,\textbf{0.85}\se{0.04} \\
LenDB & 574K & 576K & \textbf{0.05}\,/\,0.99\se{0.01} & \textbf{0.05}\,/\,\textbf{1.00}\se{0.00} \\
Seismic & 563K & 558K & 0.04\,/\,0.97\se{0.02} & \textbf{0.03}\,/\,\textbf{0.99}\se{0.01} \\
MosquitoSound & 2.95M & 2.95M & 0.10\,/\,0.91\se{0.03} & \textbf{0.06}\,/\,\textbf{0.98}\se{0.01} \\
FruitFlies & 2.95M & 2.95M & 0.10\,/\,0.94\se{0.02} & \textbf{0.06}\,/\,\textbf{0.98}\se{0.01} \\
\midrule
Average & -- & -- & 0.07\,/\,0.90\se{0.01} & \textbf{0.05}\,/\,\textbf{0.95}\se{0.01} \\
\bottomrule
\end{tabular*}
\end{table}